\documentclass[letterpaper,10pt,conference]{ieeeconf}

\IEEEoverridecommandlockouts

\usepackage[dvipsnames]{xcolor}
\usepackage{cite}
\usepackage{amsmath}
\usepackage{amssymb}
\usepackage{graphicx}
\usepackage{xspace}
\usepackage{hyperref}
\hypersetup{
    colorlinks=true,
    linkcolor=orange,
    filecolor=magenta,
    urlcolor=orange,
    citecolor=orange,
}
\usepackage[capitalise, nameinlink]{cleveref}

\newcommand{\textgbf}[1]{\textcolor{darkgray}{\textbf{#1}}}

\newcommand{\BenchName}{\textbf{RoboCasa-X}\xspace}

\title{\textbf{Cross-Embodiment Transfer via Behavior-Aligned Representations}}

\author{Ajay Sridhar$^{*1}$,
Jensen Gao$^{*1}$,
Jonathan Yang$^1$,
Jean Mercat$^2$,
Suneel Belkhale$^1$,
Dorsa Sadigh$^1$
\vspace{-20pt}
\thanks{*Equal contribution. $^{1}$Stanford University, $^{2}$Toyota Research Institute. Correspondence to: {\footnotesize\ttfamily ajaysri@stanford.edu}.}
}

\begin{document}

\maketitle
\thispagestyle{empty}
\pagestyle{empty}

\begin{abstract}
Recent progress in large-scale imitation learning for robot manipulation has been driven by leveraging datasets across a wide range of robot embodiments. However, achieving significant cross-embodiment transfer is often still challenging. In this work, we study the role of using behavior-aligned representations (e.g., object bounding boxes, language motions, end-effector traces of robot motion) in vision-language-action (VLA) models to promote cross-embodiment transfer. We hypothesize that by possessing invariances across embodiments while being predictive of robot actions, these representations can help unify large-scale cross-embodiment data to enhance transfer. To assess our hypothesis, we develop a simulation-based benchmark designed to assess transfer with diverse cross-embodiment data to new embodiments. Using this benchmark, we compare different representations and ways of incorporating them. We identify that end-effector traces can be particularly beneficial for transfer, representations are generally more useful with larger prior datasets, and can be used to benefit from action-free data. We also demonstrate that they can enhance sim-to-real cross-embodiment transfer, improving task completion progress of real robot policies pre-trained on simulation data by 28\%. We provide videos of our evaluations at our website \url{https://ajaysridhar.com/barx/}.

\end{abstract}

\section{Introduction}
\label{sec:introduction}
Modern machine learning has shown that general-purpose models trained on large, diverse datasets outperform specialized models trained on narrow, domain-specific data~\cite{achiam2023gpt, radford2021learning, kirillov2023segment}. This has inspired researchers in robot learning to scale robot datasets~\cite{khazatsky2024droid, walke2023bridgedata}, with the hope of achieving more robust and capable policies.
However, data collection efforts are usually specific to individual robot platforms. To further scale robot learning and benefit from as much data as possible, it becomes important to learn cross-embodiment policies that can leverage data from a wide variety of robot platforms and transfer abilities across embodiments and tasks~\cite{o2023open, shah2023vint, Doshi24-crossformer}.
Furthermore, cross-embodiment policies have the potential to aid transfer to entirely new embodiments, which is useful due to the constantly evolving nature of robot hardware.

However, despite the promise of cross-embodiment learning, prior work has seen mixed results in achieving successful cross-embodiment transfer for manipulation. For example, while there has been evidence that cross-embodiment data can improve various generalization axes~\cite{taxonomy2025arxiv} or help transfer skills across embodiments~\cite{o2023open}, often times performance does not significantly exceed training only on data specific to the target embodiment~\cite{o2023open, Doshi24-crossformer}.
This is most likely due to substantial variation between policy inputs (e.g., image observations) and outputs (e.g., action spaces) across embodiments, and not enough data coverage to overcome this.

Prior works have improved cross-embodiment transfer through explicitly aligning observations~\cite{yang2023polybot, chen2024mirage, chen2024rovi, lepert2024shadow} or action spaces~\cite{yang2023polybot, yang2024pushinglimitscrossembodimentlearning, zheng2025universalactionsenhancedembodied}. However, these works have various limitations, such as requiring significant manual effort (e.g., aligning camera poses), or knowledge of the deployment-time robot during training, which can make them challenging to scale.
In order to better promote cross-embodiment transfer at scale, we need methods for aligning heterogeneous datasets that are more scalable and general.

Our insight is that while observation and action spaces are often difficult to explicitly align, we can instead \emph{implicitly} align them. In particular, we can connect data from different embodiments through \emph{behavior-aligned} representations -- representations that contain information relevant for action prediction --
such as language descriptions~\cite{belkhale2024rthactionhierarchiesusing} or 2D visual traces of robot motion~\cite{gu2023rttrajectory, niu2024llarva}.
While prior works have shown the utility of these representations for robot learning, their role in promoting cross-embodiment transfer has not been studied in detail.
We hypothesize these representations can offer an additional space for policy reasoning that is invariant across embodiments, enabling scalable transfer.

To test this hypothesis, we conduct an empirical study on how different behavior-aligned representations affect cross-embodiment transfer. In particular, we consider training vision-language-action (VLA) models to predict these representations jointly with actions, and vary different representations and ways of incorporating them. To facilitate our study, we develop \BenchName, a simulated benchmark based on prior work~\cite{robocasa2024}, which involves pre-training policies using diverse, cross-embodiment data, and then assessing transfer to new embodiments with limited additional data.

Using our benchmark, we find that a variety of representations can enhance transfer, but traces of end-effector motion are generally the most impactful out of those we consider. Furthermore, we show that representations scale well with larger prior datasets, predicting representations during inference is not necessary for transfer, and representations can facilitate transfer from action-free datasets. Lastly, we evaluate on real robots, where we use data from \BenchName for sim-to-real cross-embodiment transfer. We find that representations can improve task progress of policies pre-trained with simulation data by 28\%. We provide videos of our evaluations at our website \url{https://ajaysridhar.com/barx/}.

\begin{figure*}
    \centering
    \includegraphics[width=1.\linewidth]{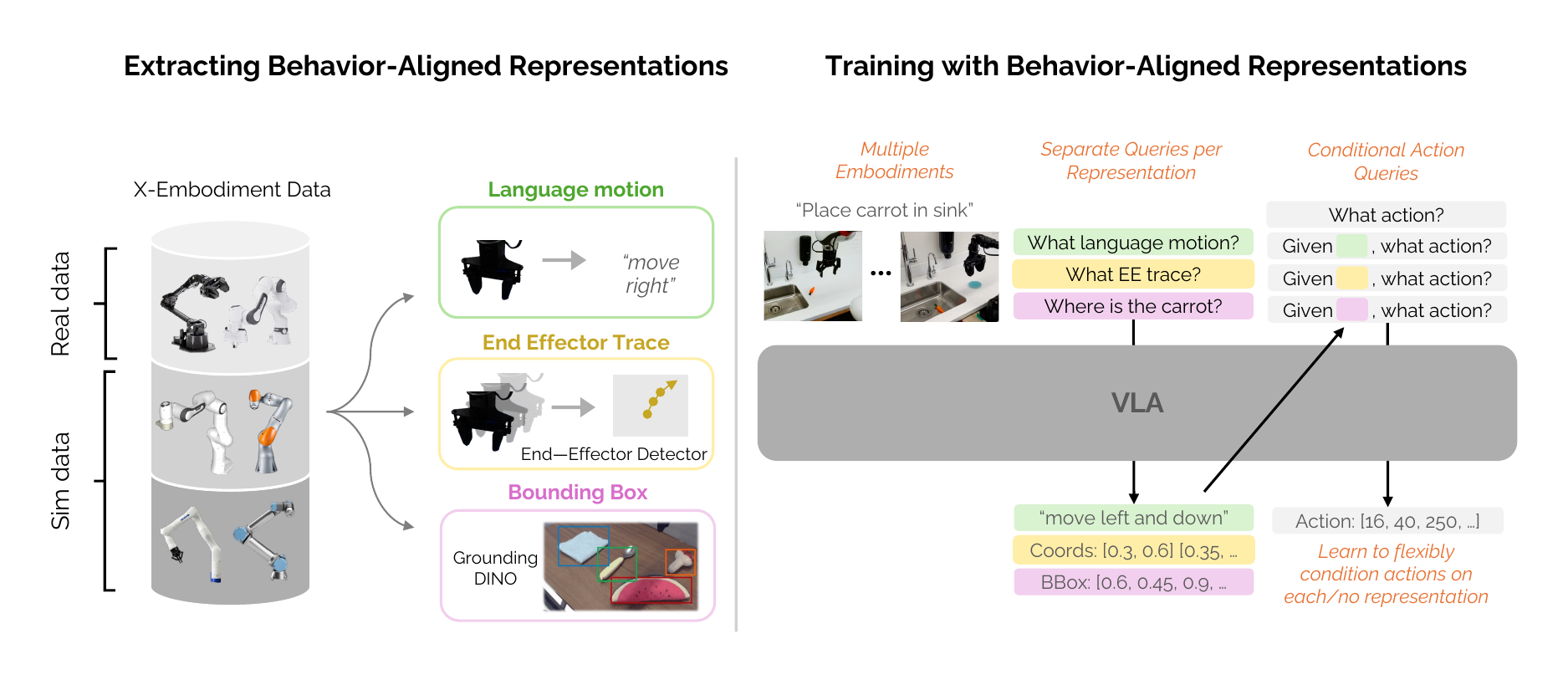}
    \vspace{-25pt}
    \caption{\small \textbf{Left}: We consider cross-embodiment datasets of both simulated and real-world robots. We then extract behavior-aligned representations: \emph{language motions} via heuristics and end effector pose deltas, \emph{end effector} traces via segmentation models, and \emph{bounding boxes} via Grounding DINO. \textbf{Right}: We train VLAs with different choices of representations and ways of incorporating them, and study how this impacts transfer from cross-embodiment datasets.}
    \vspace{-10pt}
    \label{fig:method}
\end{figure*}

\section{Related Work}
\label{sec:related_works}
In this section, we review prior work on cross-embodiment learning, vision-language-action models, and using auxiliary objectives and representations for robot learning.

\smallskip \noindent \textbf{Cross-Embodiment Learning.}
Cross-embodiment learning has emerged as a popular paradigm for scaling robot manipulation data. However, differences between embodiments in their observations and actions can pose a significant challenge for transfer. Some efforts to leverage cross-embodiment data train a single model with minimal data alignment, which can achieve some degree of transfer~\cite{o2023open, octo_2023, kim24openvla, black2024pi_0, team2025gemini}. Others use a shared model backbone with specialized action heads or controllers for different embodiments~\cite{Doshi24-crossformer, yang2023polybot}. Other works align observations, such as by using wrist cameras, inpainting, or segmentation masks~\cite{yang2023polybot, chen2024mirage, chen2024rovi, lepert2024shadow, yang2024pushinglimitscrossembodimentlearning}. However, these approaches can be difficult to scale. Rather than explicitly aligning observations or actions, we study using \emph{behavior-aligned} representations (e.g., language motions, end-effector traces), which we hypothesize can also help unify cross-embodiment data, but in a more implicit, scalable manner.

\smallskip \noindent \textbf{Vision-Language-Action Models.}
Vision-language-action (VLA) models have become an attractive policy class for generalist robot manipulation~\cite{brohan2023rt, kim24openvla, belkhale2024minivla, black2024pi_0, szot2024multimodal, team2025gemini, intelligence2025pi}. These policies involve fine-tuning vision-language models (VLMs) to predict robot actions, such that the policy can inherit knowledge from internet-scale vision-language data to enhance generalization. Their high-capacity architectures also allow them to effectively ingest large and heterogeneous robot datasets. In our work, we use VLAs as our policies due to these appealing qualities. Furthermore, because VLAs can also be trained to predict text~\cite{brohan2022rt, belkhale2024rthactionhierarchiesusing, nasiriany2024rta, Zawalski24ecot}, we use this as a simple way of incorporating behavior-aligned representations.

\smallskip \noindent \textbf{Auxiliary Representations for Robot Learning.}
Several works have used auxiliary training objectives to improve policy performance and robustness. Co-training VLA policies on visual question answering (VQA) data has been shown to benefit generalization to novel task semantics and scene visuals~\cite{taxonomy2025arxiv, brohan2023rt, intelligence2025pi}. Other works use intermediate policy representations to guide action prediction. These representations include image points~\cite{sundaresan2023kite, wen2023any}, end effector traces~\cite{niu2024llarva, gu2023rttrajectory,wang2023mimicplay} and poses~\cite{nasiriany2024rta}, language motion descriptions~\cite{belkhale2024rthactionhierarchiesusing}, goal images~\cite{black2023zero, sundaresan2024rt, zhao2025cot}, object bounding boxes~\cite{minderer2022simpleopenvocabularyobjectdetection, intelligence2025pi, team2025gemini}, and segmentation masks~\cite{huang2025robogroundroboticmanipulationgrounded}. These representations can also be combined in a chain-of-thought manner~\cite{Zawalski24ecot, clark2025action, Chen25-ecot-lite}. While these works demonstrate the general utility of these representations, we focus on studying their effect on cross-embodiment transfer. Some works have used these representations with cross-embodiment data~\cite{wen2023any, niu2024llarva, intelligence2025pi}, but our work analyzes different representations, ways of incorporating them, and more carefully analyzes their effect on cross-embodiment transfer with diverse tasks.

\section{Behavior-Aligned Representations for Cross-Embodiment Transfer}
\label{sec:method}
In this section, we first overview our framework for learning policies with behavior-aligned representations. Next, we discuss our instantiation of this framework, including the specific representations we consider, our annotation process for obtaining them, and other details.

\subsection{Formalization}

We consider the language-conditioned cross-embodiment imitation learning (IL) setting. We define a robot embodiment space $\mathcal{R}$ where $r \in \mathcal{R}$ designates a specific embodiment, which has an associated observation space $\mathcal{O}^r$ and action space $\mathcal{A}^r$. These spaces may have some relation/alignment across robots, but we do not assume this. Our goal is to learn a policy \(\pi_\theta(a \mid o, l)\) that maps observations $o \in \mathcal{O}^r$ and language instructions \(l \in \mathcal{L}\) to actions $a \in \mathcal{A}^r$. We assume expert data of the form \(\mathcal{D} = \{(o^r_i, a^r_i, l_i)\}_{i=1}^N\), where $r$ designates which embodiment a given observation or action is from. In behavior cloning, the policy is trained to minimize a supervised loss function $\ell$ to imitate expert actions:
\[
\mathcal{L}_{\text{BC}}(\theta) = \mathbb{E}_{(o, a, l) \sim \mathcal{D}} \left[ \ell(\pi_\theta(\cdot \mid o, l), a) \right].
\]
In addition, we assume access to a set of \emph{behavior-aligned} representations. These can be any quantity that can be computed from $\mathcal{D}$, contains information useful for predicting optimal behavior, and has invariance to the embodiments in $\mathcal{R}$, meaning that different embodiments should have similar representations for the same task. Each observation \(o_i\) is annotated with a tuple of representations \(z_i = (z_i^{(1)}, \dots, z_i^{(K)})\), where each \(z_i^{(k)} \in \mathcal{Z}^{(k)}\) belongs to one of $K$ different representation spaces.

To leverage these representations, we modify the policy $\pi_\theta$ to condition on subsets of them when predicting actions.
We also train $\pi_\theta$ to predict these representations.
For each training example, we sample a subset of representations, which we denote $\tilde{z} \subseteq \{z^{(1)}, \dots, z^{(K)}\}$, using some pre-defined representation distribution $p_\text{rep}$, i.e., $\tilde{z}_i \sim p_\text{rep}(\tilde{z})$.
Our total loss combines the behavior cloning objective with auxiliary supervision on the representations:
\begin{equation*}
    \mathcal{L}_{\text{total}}(\theta) = \mathbb{E}_{(o,z,a,l) \sim \mathcal{D}, \tilde{z} \sim p_\text{rep}(z)} \left[\ell(\pi_\theta(\cdot \mid o, l, \tilde{z}), a) + \ell_{\text{aux}} \right],
\end{equation*}
\begin{equation*}
     \ell_{\text{aux}} =  \sum_{k=1}^K \lambda_k \ell^{(k)}_{\text{rep}}(\pi_\theta(\cdot \mid o, l), z^{(k)}).
\end{equation*}

Here, $\ell^{(k)}_{\text{rep}}$ is the loss for predicting the $k$th representation, weighted by $\lambda_k$. In practice, we instantiate $\pi_\theta$ as a VLA. Then, our formulation amounts to autoregressively predicting a subset of behavior-aligned representations as text, and then actions, as a single sequence. The model receives different text prompts to indicate which subset of representations it should predict. This conditioning scheme is similar to prior works that predict representations as forms of embodied reasoning before predicting actions~\cite{belkhale2024rthactionhierarchiesusing,Zawalski24ecot, Chen25-ecot-lite}.

We note that this approach can be applied to any dataset of language-annotated demonstrations, but we consider it specifically for cross-embodiment learning. We hypothesize that because these representations have invariance across embodiments and are useful for action prediction, they can aid cross-embodiment transfer through implicit alignment.

\subsection{Choice of Representations}
\label{sec:representations}
We describe the specific representations we study and our methods for annotating them. Drawing from prior work, we choose a set of representations that are simple to annotate, informative for predicting actions, and have inherent invariance across different embodiments. We note that the best choice of representations may vary, depending on which have invariance to the robot platforms for a particular setting.

\smallskip \noindent \textbf{Bounding Boxes}. We use bounding boxes for the current image observation of objects that the robot should manipulate \cite{stone2023openworldobjectmanipulationusing, Zawalski24ecot, intelligence2025pi}. To label our real-world datasets, we use an off-the-shelf pipeline from prior work \cite{Zawalski24ecot} consisting of VLM scene descriptions and Grounding DINO~\cite{liu2024groundingdinomarryingdino}.

\smallskip \noindent \textbf{Language Motions}: We use language descriptions of the motion corresponding to the action the robot should execute next, such as ``move left and down''. We obtain this representation through a pipeline similar to prior work \cite{belkhale2024rthactionhierarchiesusing}, based on thresholding changes in robot proprioceptive state.

\smallskip \noindent \textbf{End Effector Traces.} We use traces of the robot end effector position during the future motion it should execute for its task \cite{gu2023rttrajectory, niu2024llarva}. These traces consist of sequences of future 2D positions of where the end effector is in the image observation frame. We use a pre-trained model from prior work~\cite{niu2024llarva} to detect these positions from image observations.

\subsection{Implementation Details}
We use the MiniVLA architecture~\cite{belkhale2024minivla} for our policies, starting from its pre-trained VLM, but without any robot pre-training. We use single third-person camera views as our observation space. We set all loss weights $\lambda_k = 1$.

\section{Simulation Experiments}
\label{sec:sim_experiments}
\begin{figure*}[t]
    \centering
    \vspace{3pt}
    \includegraphics[width=0.9\linewidth]{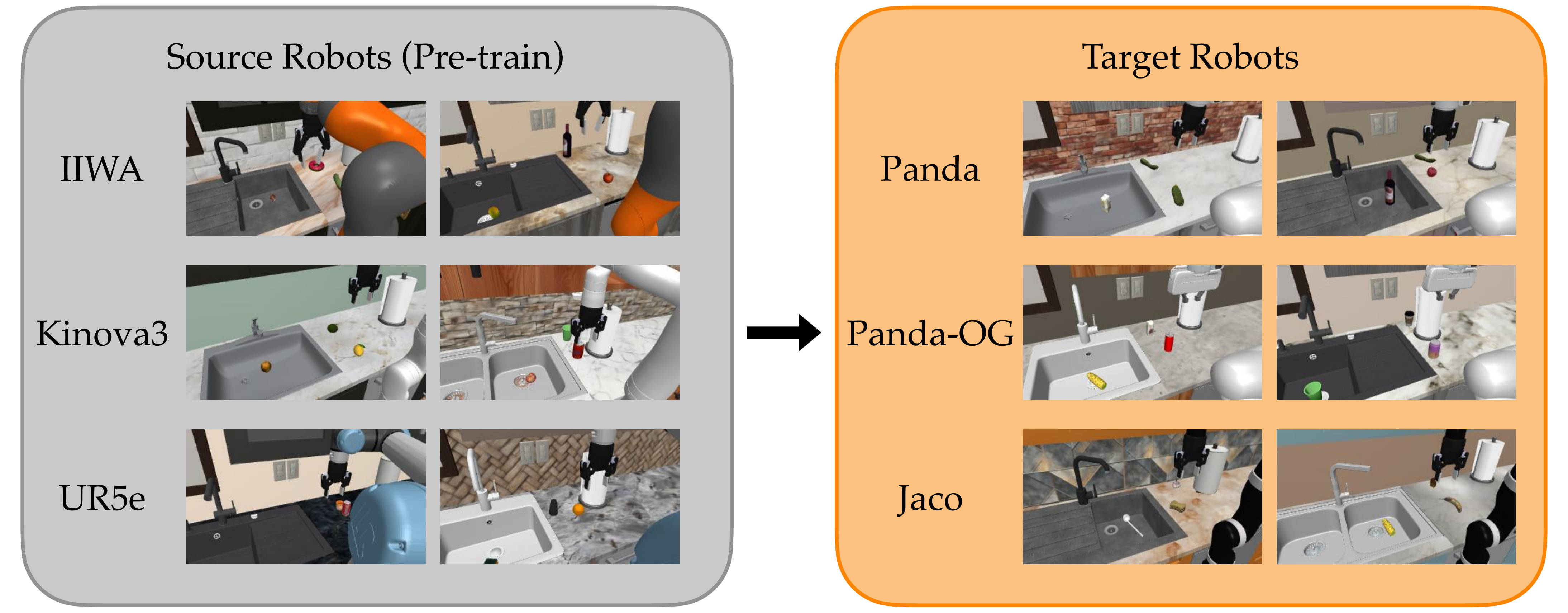}
    \caption{\small We pre-train policies on diverse cross-embodiment data mixtures from three source robots (left), and then transfer these policies to different target robots using limited data (right). We show initial observations in our data for the task \emph{PnP Counter to Sink}. Our tasks capture significant variation in kitchen layouts, textures, object types, and object poses.}
    \label{fig:data_examples}
\end{figure*}

\begin{figure*}[t]
    \centering
    \includegraphics[scale=0.3]{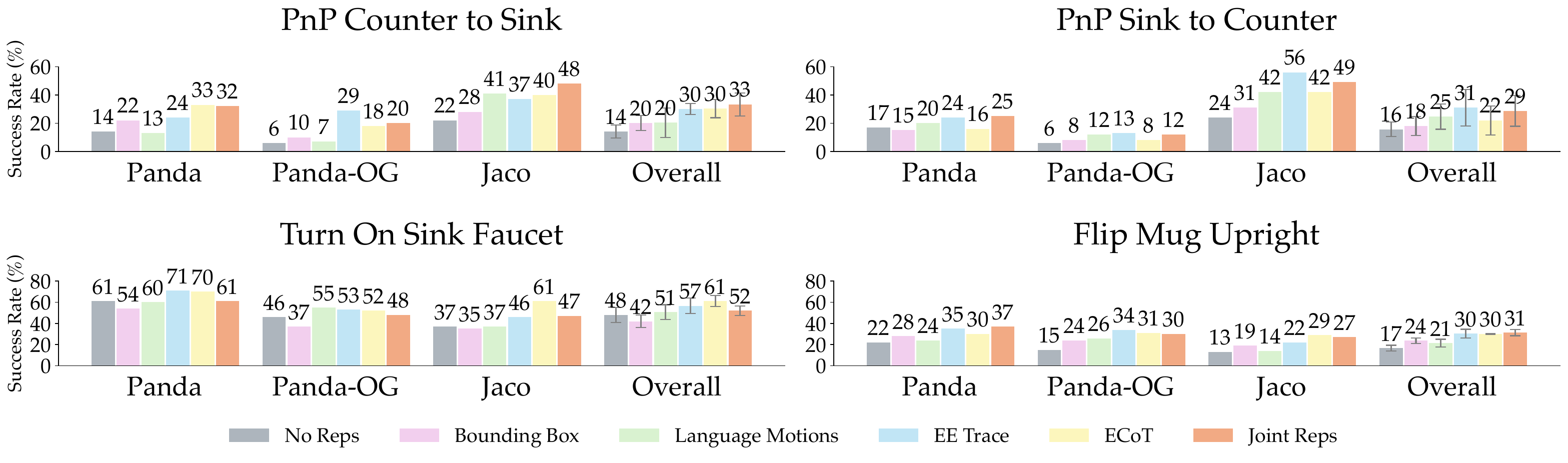}
    \vspace{-5pt}
    \caption{\small We compare training with no representations, each representation separately, and combined using either \textbf{ECoT} or \textbf{Joint Reps}. Each representation improves transfer over using none, with end-effector traces helping the most. \textbf{Joint Reps} performs the best overall, except on Panda-OG, where end-effector traces alone are better.}
    \label{fig:train_reps}
    \vspace{-10pt}
\end{figure*}

\subsection{Benchmark Design}
Although prior work has studied cross-embodiment learning for manipulation using diverse data in the real world~\cite{o2023open, octo_2023, Doshi24-crossformer, kim24openvla}, to our knowledge, simulation benchmarks for this setting have not previously been considered. Therefore, we design a new simulation benchmark for cross-embodiment learning called \BenchName, based on RoboCasa~\cite{robocasa2024}, a platform that supports realistic and diverse scenes and tasks. We consider variations of three tasks from RoboCasa (\emph{PnP Counter to Sink}, \emph{PnP Sink to Counter}, \emph{Turn On Sink Faucet}), as well as a new task (\emph{Flip Mug Upright}). These tasks capture a large amount of variation, including different kitchen layouts, textures, object types, and poses, making them challenging and suitable for studying transfer.

To create large-scale prior datasets, we use MimicGen \cite{mandlekar2023mimicgen} to generate data for three source robot embodiments: IIWA, Kinova3, and UR5e. The Kinova3 and UR5e use the Robotiq 2F-85 gripper, while the IIWA uses the Robotiq 2F-140. To simulate realistic diversity and promote sim-to-real transfer (as done later in \cref{sec:real_experiments}), we randomize the camera pose. We vary dataset size with 300 or 1000 demonstrations per task/embodiment, for a total of 900 or 3000 demonstrations per task. We refer to these datasets as \textgbf{XP-900} and \textgbf{XP-3K} (short for Cross Prior), respectively.

We use three additional embodiments as target robots: Panda and Jaco with the Robotiq 2F-85, and Panda with its original Franka Hand gripper (designated as Panda-OG). For each target robot, we collect 50 human demonstrations per task.
We measure cross-embodiment transfer by pre-training on the prior robot datasets, and then transferring to each target robot by co-fine-tuning on the target demonstrations and prior data. Because 50 target demonstrations are not enough to capture the diversity in each task, policy performance is reflective of cross-embodiment transfer from the prior data.

In \cref{fig:data_examples}, we visualize initial observations for the source embodiments in the pre-training data for the task \emph{PnP Counter to Sink}, and compare this to the initial observations for the target robots. Our other tasks are instantiated in similar scenes with similar types of variation.

\smallskip \noindent \textbf{Additional Details.} Because we have access to privileged information in simulation, we obtain ground truth labels for bounding boxes and end effector traces, rather than the annotation methods described in \cref{sec:representations}.
For each evaluation, we conduct 100 rollouts with a fixed set of scene conditions. We report the highest success rate out of three checkpoints for each model. We evaluate using the same camera view for all robots (same as in the target robot data). All of our simulated robots share the same action space (delta Cartesian end effector pose control) and controller.

\subsection{Experiments}

\noindent In our experiments, we aim to address the following:

\begin{enumerate}
    \renewcommand{\labelenumi}{\textbf{Q\arabic{enumi}}}
    \item \textbf{(Incorporating Representations)}: Which representations help most with transfer to new embodiments, and how should they be incorporated?
    \item \textbf{(Cross-Embodiment Scaling)}: Are representations more helpful with larger, cross-embodiment datasets than smaller, single-embodiment datasets?
    \item \textbf{(Representation Inference)}: Does predicting and conditioning on representations matter during inference?
    \item \textbf{(Action-Free Transfer)}: Can representations help with cross-embodiment transfer from action-free data?
\end{enumerate}

\begin{figure*}[th]
    \centering
    \vspace{3pt}
    \includegraphics[scale=0.3]{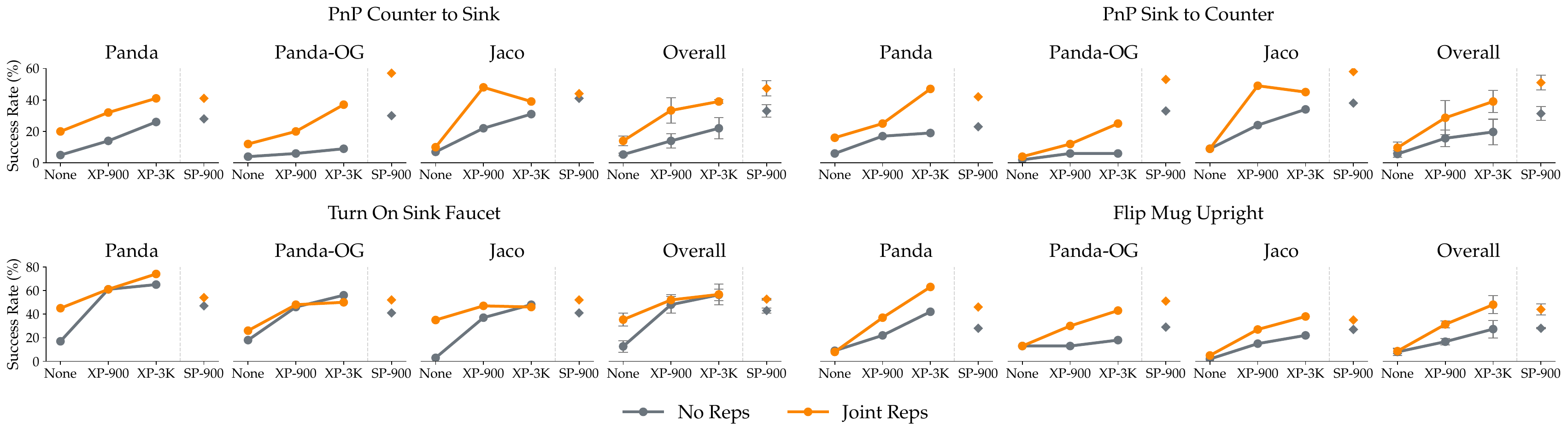}
    \vspace{-5pt}
    \caption{\small We compare the effect of behavior-aligned representations when using different scales of prior data. Representations are generally more impactful when scaling up prior data (with the exception for the task \emph{Turn On Sink Faucet}).}
    \label{fig:adaptation}
\end{figure*}

\begin{figure*}[t]
    \centering
    \includegraphics[scale=0.28]{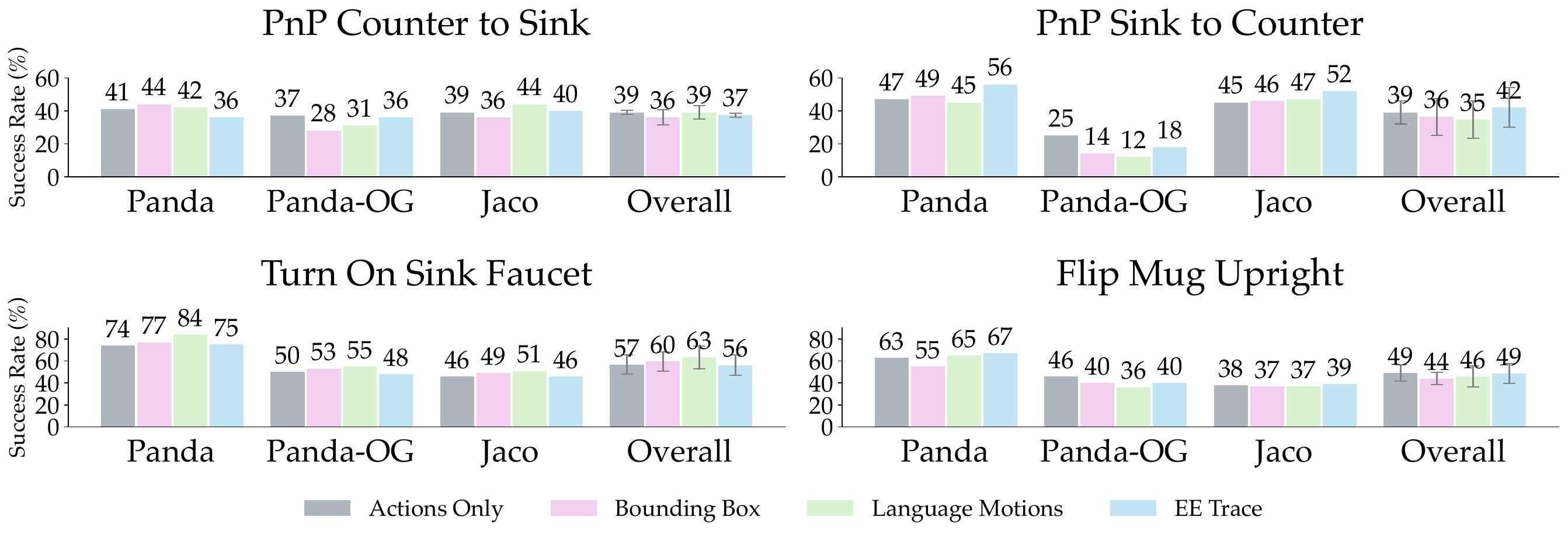}
    \caption{\small We compare inference of \textbf{Joint Reps} models by either only predicting actions, or first predicting a representation and then actions. Predicting representations can help in some situations, but the impact is usually not substantial.}
    \label{fig:eval_reps}
    \vspace{-10pt}
\end{figure*}

\smallskip \noindent \textbf{Incorporating Representations.} First, we analyze how each representation contributes to cross-embodiment transfer, and how to incorporate them during training. We consider the following approaches, which amount to different choices of representation distributions $p_\text{rep}(\tilde{z})$ in our framework:
\begin{enumerate}
    \item \textbf{No Reps}: We train the model to only predict actions, without any representations.
    \item \textbf{Single Rep}: We train the model to either predict a single representation and then actions, or only actions. We train separate models for each representation.
    \item \textbf{ECoT}~\cite{Zawalski24ecot}: We train the model to either predict each representation and then actions as a sequential chain (bounding boxes $\rightarrow$ end-effector trace $\rightarrow$ language motion $\rightarrow$ actions), or only actions.
    \item \textbf{Joint Reps}: Similar to \textbf{Single Rep}, except we train a single model with all representations (e.g., the model is trained to either predict one of multiple representations and then actions, or only actions).
\end{enumerate}

\noindent For each approach, we use it to pre-train a model on \textgbf{XP-900}. Then, separately for each target robot, we co-fine-tune it with the addition of target robot data. We do this procedure with separate models for each task, except for the two \emph{PnP} tasks, where we use the same model for both tasks. During inference for each model, we only predict actions without representations (which each model was also trained to do), as we found this to significantly speed up inference without significantly affecting performance.

In our results in \cref{fig:train_reps}, we find that using each representation individually improves transfer overall compared to using none. End-effector traces are generally the most helpful, followed by language motions, and lastly bounding boxes. When combining representations using either \textbf{ECoT} or \textbf{Joint Reps}, performance generally improves over each one in isolation for Panda and Jaco with the Robotiq 2F-85.

However, combining representations does not perform as well as only using end-effector traces for Panda-OG, whose end-effector is unseen in the prior data (shown in \cref{fig:data_examples}). We hypothesize this is because traces are particularly important for aligning data with more varied end-effectors. More generally, this suggests that the best choice of representations may depend on the differences between embodiments.

\begin{figure*}[t]
    \centering
    \vspace{4pt}
    \emph{PnP Counter to Sink} \par\vspace{0.5em}

    \begin{minipage}{0.32\textwidth}
        \centering
        \includegraphics[width=\linewidth]{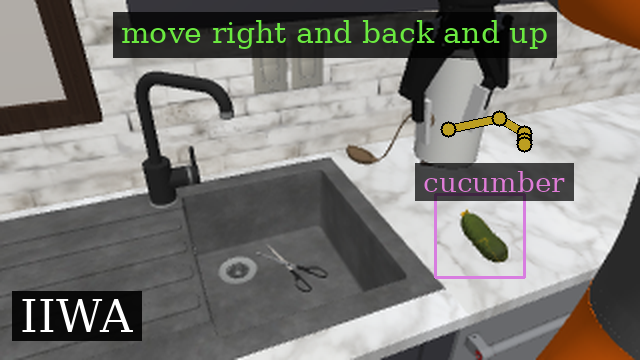}
    \end{minipage}
    \hfill
    \begin{minipage}{0.32\textwidth}
        \centering
        \includegraphics[width=\linewidth]{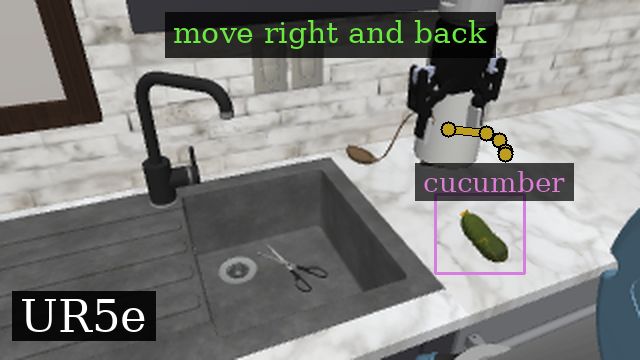}
    \end{minipage}
    \hfill
    \begin{minipage}{0.32\textwidth}
        \centering
        \includegraphics[width=\linewidth]{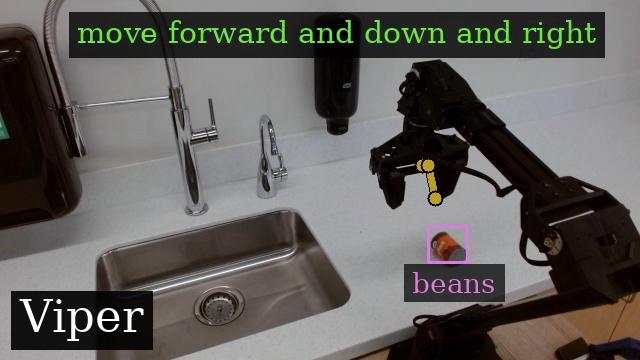}
    \end{minipage}

    \vspace{0.5em}

    \emph{PnP Sink to Counter} \par\vspace{0.5em}

    \begin{minipage}{0.32\textwidth}
        \centering
        \includegraphics[width=\linewidth]{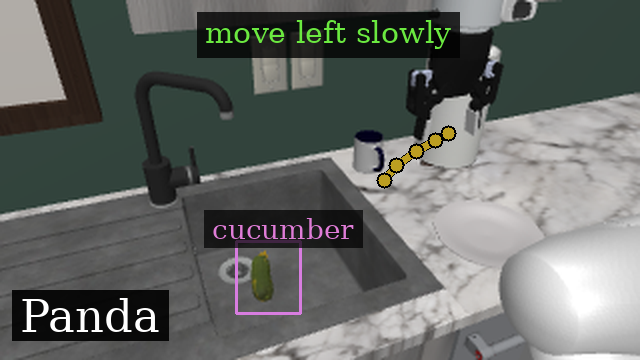}
    \end{minipage}
    \hfill
    \begin{minipage}{0.32\textwidth}
        \centering
        \includegraphics[width=\linewidth]{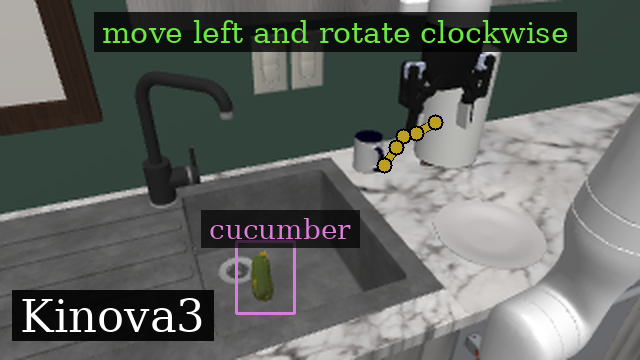}
    \end{minipage}
    \hfill
    \begin{minipage}{0.32\textwidth}
        \centering
        \includegraphics[width=\linewidth]{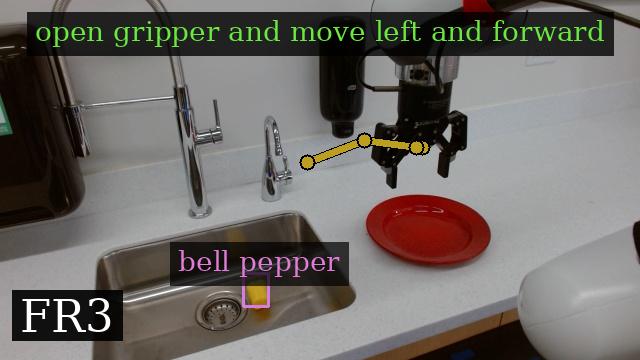}
    \end{minipage}
    \caption{\small We show examples of our real tasks (right) in comparison to their sim counterparts (left, middle). We also show actual predictions from \textbf{Joint Reps} models, which demonstrate similar predicted representations across different embodiments in unseen scenes.}
    \label{fig:real_alignment}
    \vspace{-10pt}
\end{figure*}

\smallskip \noindent \textbf{Cross-Embodiment Scaling.}
Next, we investigate if behavior-aligned representations are more beneficial with larger cross-embodiment datasets than smaller single-embodiment datasets.
To do this, we compare \textbf{No Reps} and \textbf{Joint Reps}. We train using the same pre-training and co-fine-tuning procedure as before, but vary the prior dataset as either \textgbf{XP-900} and \textgbf{XP-3K}. We also compare to training on the target robot data from scratch with no prior (\textgbf{None}), and using a prior dataset created in the same manner as \textgbf{XP-900}, but with the same robot as the target (\textgbf{SP-900}).

In \cref{fig:adaptation}, we find that behavior-aligned representations help when training only on target robot data, either with no prior or the same embodiment prior, corroborating prior work~\cite{niu2024llarva, Chen25-ecot-lite}. However, for all of the tasks but one (\emph{Turn On Sink Faucet}), the overall improvement when training with no prior (+5\%) is smaller than when using \textgbf{XP-900} (+15\%) or \textgbf{XP-3K} (+19\%). This suggests that representations scale effectively with larger cross-embodiment datasets.

On Panda and Jaco, training on \textgbf{XP-3K} with representations often matches or even exceeds training on the same embodiment prior (\textgbf{SP-900}), suggesting that the degree of transfer is strong. However, this is not the case with Panda-OG, where performance lags significantly behind training on the same embodiment prior, suggesting that there is still room for improvement with larger embodiment gaps.

Unlike the other tasks, for \emph{Turn On Sink Faucet}, representations have a larger impact when training only on target robot data (+22\%) than when using \textgbf{XP-900} (+4\%) or \textgbf{XP-3K} (+1\%). We hypothesize this is because \emph{Turn On Sink Faucet} has less variation than our other tasks, which involve manipulating a variety of objects in different poses, while there is not as much variation in the poses and types of faucets. Therefore, performance is saturated more easily, and prior data is not as critical.

\smallskip \noindent \textbf{Representation Inference.}
So far, we have evaluated all models by directly predicting actions, rather than first predicting representations and then actions. Concurrent work~\cite{Chen25-ecot-lite} has found that much of the benefit from representations can be obtained during training alone, which is desirable to reduce the latency introduced by predicting representations. We investigate this further in the cross-embodiment setting by evaluating \textbf{Joint Reps} models trained using \textgbf{XP-3K} with each supported inference method (i.e., predicting actions only, or predicting one representation and then actions). In \cref{fig:eval_reps}, we find that predicting representations can help slightly in some settings, although the overall effect is not substantial. This corroborates the findings from~\cite{Chen25-ecot-lite}, and suggests that representations primarily enhance cross-embodiment transfer in an implicit manner.

\begin{figure}[h]
    \centering
    \includegraphics[scale=0.19]{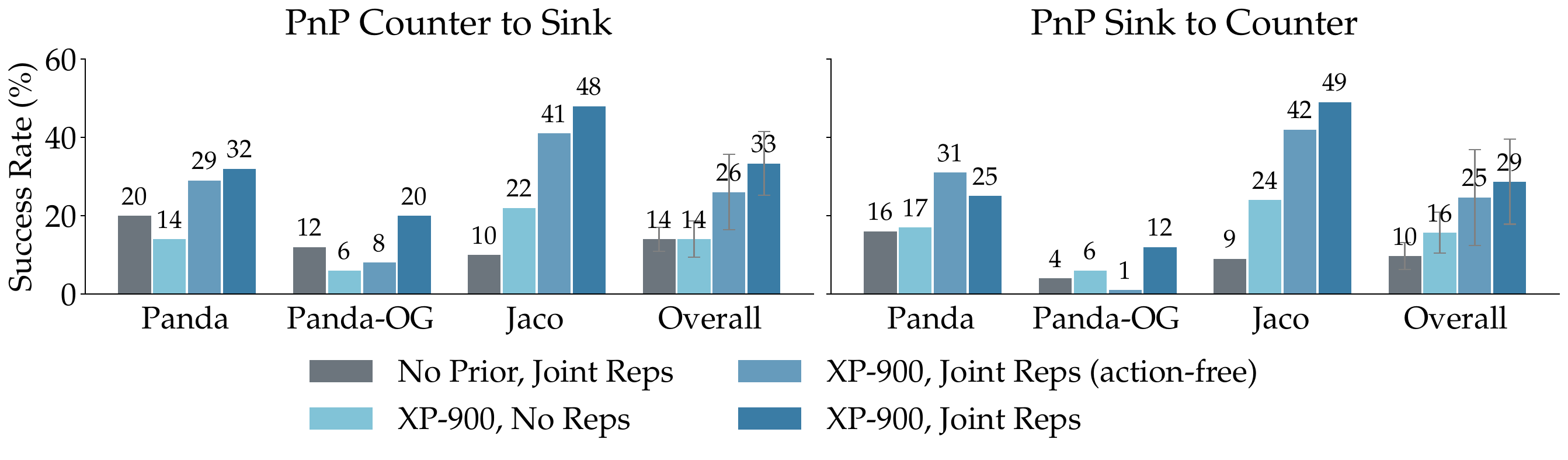}
    \vspace{-10pt}
    \caption{\small We find that behavior-aligned representations can leverage action-free prior datasets to improve over learning with no prior dataset, as well as using the full prior dataset with actions but without representations. However, performance is slightly worse than using representations with the prior dataset with actions.}
    \label{fig:action_free}
    \vspace{-5pt}
\end{figure}

\smallskip \noindent \textbf{Action-Free Transfer.}
We conduct additional experiments to assess if representations can improve cross-embodiment transfer from action-free datasets. To do this, we pre-train a variant of our \textbf{Joint Reps} model for our two \emph{PnP} tasks on \textgbf{XP-900}, but without action prediction (only representations). Then, we co-fine-tune on target robot data, but with both representations and action prediction on the target robot data. This is similar to \emph{reasoning pre-training} as proposed in~\cite{Chen25-ecot-lite}. Because our annotations for language motions are obtained using actions/proprioception, we do not include this representation when training on our action-free prior dataset.

In our results in \cref{fig:action_free}, we find that action-free pre-training with representations does generally help, improving by 14\% overall compared to learning from scratch with the target robot data, and by 11\% compared to using the full prior dataset with actions, but without representations. There is a performance decrease compared to using the prior dataset with actions, but much of the performance is retained, except for with Panda-OG, where action-free pre-training does not help. This suggests that behavior-aligned representations can effectively induce transfer from action-free cross-embodiment data through the internal representations they induce, although this may depend on how similar the target embodiment is to those in the action-free data.

\section{Real-World Experiments}
\label{sec:real_experiments}
To see if behavior-aligned representations also enhance real-world cross-embodiment learning, we consider two real target embodiments: the Franka Research 3 (FR3) and ViperX 300 S. We evaluate on variations of the \emph{PnP} tasks in \BenchName. We adapt our models pre-trained on \textgbf{XP-3K} using 50 demonstrations per task. Our setting represents a similar paradigm for sim-to-real transfer as prior work \cite{maddukuri2025simandreal}, but also considers cross-embodiment transfer, as the real target robots are not in the simulation data. We loosely align the camera view and environment layout from simulation, but there remain significant domain gaps. We compare examples of our real tasks to their simulation counterparts in \cref{fig:real_alignment}.

\smallskip \noindent \textbf{Action Alignment.} Unlike in our simulation experiments, there are now also significant differences in the action spaces between the source and target robots. Although all robots use delta end effector pose control, their control stacks differ significantly, including differences in control frequency, and blocking versus non-blocking control. This makes transfer in this setting especially challenging.

\smallskip \noindent \textbf{Experimental Setup.} For practicality of evaluation, we modify the real-world instantiations of the \BenchName tasks to have less variation. For instance, we only use one kitchen, and reduce the number of object types to four for \emph{PnP Counter to Sink} and one for \emph{PnP Sink to Counter}.
Like in our simulation experiments, we train models with and without cross-embodiment prior data (\textgbf{XP-3K}), using \textbf{No Reps} and \textbf{Joint Reps}. Unlike our simulation experiments, for our models with prior data, we train a single model on both target embodiments, to allow for possible transfer.
For each embodiment/task pair, we evaluate for 10-30 rollouts. We report task completion progress of each model.

\begin{figure}[h]
    \centering
    \includegraphics[scale=0.21]{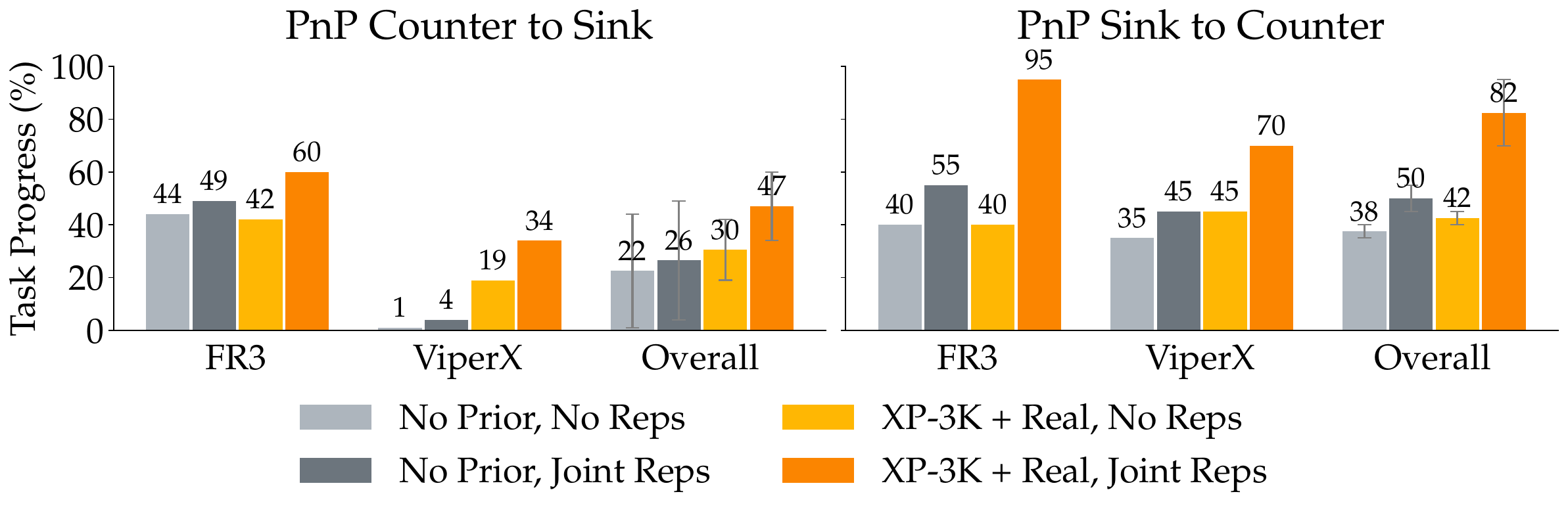}
    \caption{\small We adapt models trained on cross-embodiment prior data from simulation to real-world target robots. Behavior-aligned representations significantly improve transfer from prior datasets despite gaps in visual observations and action spaces.}
    \label{fig:real_adapt}
    \vspace{-5pt}
\end{figure}

\smallskip \noindent \textbf{Results.}
In \cref{fig:real_adapt}, we see that \textbf{Joint Reps} significantly improves task progress with cross-embodiment data (orange bars), with an overall increase of 28\%. Furthermore, representations help more when training with the cross-embodiment data than when only using target robot data (+8\%, gray bars), corroborating our simulation results. Without representations, the cross-embodiment data provides limited benefit (+7\%) over using the target data alone, suggesting that the representations help enable cross-embodiment transfer that otherwise is difficult to achieve.

The benefit of behavior-aligned representations is more pronounced in our real experiments than in simulation. We suspect this is due to the larger domain gap in observations and actions between the simulated and real robots, which makes cross-embodiment transfer without representations more challenging. In \cref{fig:real_alignment}, we show examples of how \textbf{Joint Reps} models can predict similar representations across different embodiments in unseen scenes.

\section{Conclusion}
\label{sec:conclusion}
We study how behavior-aligned representations can enhance cross-embodiment transfer through implicit data alignment. These representations can easily be labeled with off-the-shelf models or simple transforms using action information. We demonstrate that representations consistently boost performance, with more significant gains with larger cross-embodiment datasets. Through our analysis using \BenchName, we find that end-effector traces are the most impactful representation we consider, direct action prediction performs comparably to predicting and conditioning on representations, and representations can facilitate action-free cross-embodiment transfer. Finally, we demonstrate that representations can enhance sim-to-real cross-embodiment transfer on two real robot platforms. We believe that our work is an important step towards understanding positive transfer in cross-embodiment learning.

\smallskip \noindent \textbf{Limitations.} Our work demonstrates the value of using multiple behavior-aligned representations when learning from cross-embodiment data.
However, while we were able to achieve improved cross-embodiment transfer with our approach, our setting involves significant alignment between the different embodiments we consider, especially in simulation (e.g., each embodiment in \textbf{RoboCasa-X} has the same distribution of camera poses, scenes, and tasks). While we do this to isolate the problem of cross-embodiment transfer, this also causes the representations we study to be more aligned across embodiments. Future work can investigate achieving transfer in less structured settings with greater misalignment between embodiments, and study how different levels of misalignment influence the ability for representations to facilitate transfer.

Additionally, the behavior-aligned representations we consider are not exhaustive, and these representations favor object-centric manipulation tasks. We hypothesize that other behavior-aligned representations can be incorporated that may be more beneficial depending on the tasks and embodiments involved. We hope our framework serves as a strong starting point for exploring more behavior-aligned representations to facilitate cross-embodiment generalization.

\section*{Acknowledgments}
This work was supported by ONR awards N00014-25-1-2479 and N00014-22-1-2293,
and NSF award 1941722. Ajay Sridhar is supported by the NSF Graduate Research
Fellowship. Toyota Research Institute (``TRI'') provided computational
resources to support this work. We thank Blake Wulfe, Kyle Hatch, Siddharth
Karamcheti, and Sedrick Keh for helpful discussions.

\bibliographystyle{IEEEtran}
\bibliography{references}

\end{document}